\documentclass{sig-alternate}

\begin{document}
%
% --- Author Metadata here ---
%\conferenceinfo{WOODSTOCK}{'97 El Paso, Texas USA}
%\CopyrightYear{2007} % Allows default copyright year (20XX) to be over-ridden - IF NEED BE.
%\crdata{0-12345-67-8/90/01}  % Allows default copyright data (0-89791-88-6/97/05) to be over-ridden - IF NEED BE.
% --- End of Author Metadata ---

\title{Transfer Learning for Content-Based \
Recommender Systems using Tree Matching}

\numberofauthors{3}

\author{
% 1st. author
\alignauthor{
Naseem Biadsy\\
       \affaddr{Deutsche Telekom Labs at\\
       Ben-Gurion University\\
       Beer-Sheva 84105, Israel}\\
       \email{naseem@cs.bgu.ac.il}}
% 2nd. author
\alignauthor
Lior Rokach\\
       \affaddr{Department of Information System Engineering\\
       Ben-Gurion University\\
       Beer-Sheva 84105, Israel}\\
              \email{liorrk@bgu.ac.il}
% 3rd. author
\alignauthor
Armin Shmilovici\\
       \affaddr{Department  of Information System Engineering\\
       Ben-Gurion University\\
       Beer-Sheva 84105, Israel}\\
       \email{armin@bgu.ac.il}
}

\maketitle
\begin{abstract}
In this paper we present a new approach to content-based transfer learning for solving the data sparsity problem in cases when the users' preferences in the target
domain are either scarce or unavailable, but the necessary information on the preferences exists
in another domain. We show that training a system to use such information across domains can produce better performance.
Specifically, we represent users' behavior patterns based on topological graph structures. Each behavior pattern
represents the behavior of a set of users, when the users' behavior is defined as the items they rated and
the items' rating values. In the next step we find a correlation between behavior patterns in
the source domain and behavior patterns in the target domain. This mapping is considered
a bridge between the two domains. Based on the correlation and content-attributes of the
items, we train a machine learning model to predict users' ratings in the target domain.
When we compare our approach to the popularity approach and KNN-cross-domain on a real world
dataset, the results show that on an average of 83$\%$ of the cases our approach outperforms both
methods.
\end{abstract}




\section{Introduction}
Cross domain recommenders \cite{berkovsky2007cross} aim to improve recommendation in one domain (hereafter: the target) based on knowledge in another domain (hereafter: the source). There are two approaches for cross domain: 1) those that use a mediator to construct and initialize an empty user model, or that enrich the existing user model by using provided data or a partial user model in another domain or service; 2) those that do not use a mediator.
In this paper we propose a content-based cross domain RS that uses a mediator
to suggest items to \textbf{a new user} in a target domain who has only rated items in the source domain.
We assume that: 1) the target domain has very high sparsity; 2) the source domain has low sparsity;
and 3) the new user has already rated some items in the source domain.
Based on these assumptions, traditional recommendation algorithms do not work well in the target domain for two reasons:
first, high sparsity; second, the new user and cold-start issues.
\newline
\textbf{The problem:}
Given a set of items \textit{T} in a target domain, and given a user \textit{u} who has already rated certain items in the source domain but not in the target,
what are the most $N$ preferred items in $T$ by user $u$?

We begin solving this problem by presenting a new graph-based transfer learning method for content-base recommendation.
We define the concept of behavior tree as a topological representation of users' behavior patterns.
By \textbf{\textit{users' behavior}} we mean the items rated by a user and the rating values the
user has assigned to these items.
Based on these trees, we find correlated behavior patterns in the source and the target domains which
are considered bridges between the two domains. Then, we combine the data of each bridge with the content data of the items in
a one features vector. This vector is used as a sample in the training data for a machine learning algorithm.
Here we assume that common users exist between the source and target domains.

The paper's innovation is (1) the graph structure that we employ for dealing with the problem and  (2) the items' content data
with transfer learning recommenders that we employ to improve classifier performance.
The trees structure provides a rich representation of the users' behavior. First, this structure enables items
to be clustered  according to the users' behavior, and the clusters to be represented by a forest of behavior trees.
Second, the tree's structure defines a hierarchy among the items, this hierarchy is important because it shows how much the item represents the behavior pattern it belongs to.
Finally, we find a correlation (mapping) between items in the source domain and items in the target domain, so by running a process of a tree matching
between trees in the source and trees in the target we get this mapping.
Furthermore, from the content data we gain more features to represent the
training samples for building the recommendation model.

\section{The Recommendation Framework}
\label{ch:theRecFram}
Our framework deals with the abovementioned problem in three phases: The first is described in (\ref{sub:preparing:training:data})
and it aims to preprocess the
users' data in the source and target domains. Preprocessing consists of three steps:
	1) building behavior trees for each domain;
	2) graph matching;
	3) building training samples.
The second phase (\ref{sec:model:training}) is for training a model on the training set.
The third phase, described in (\ref{sec:recommendation}), is for recommending items to a new user
based on his/her behavior in the source domain.

\subsection{Preparing Training Data}
\label{sub:preparing:training:data}
\subsubsection{Building Behavior Graphs}
\label{sub:building:behavior:graphs}
Constructing the forests is done in four steps. \textbf{First} we separate the
same item with different rating values by expanding each item into $k$ items, where $k$ equals the
number of possible rating values in the domain. For example, if item $i$ was rated as $r1$
by one group of users and the same item got rating $r2$ by a different group of users then we consider
them as different items and represent them by $i_{r1}$ and $i_{r2}$.
Note: the number of items after this step is $k \times number$  $of$ $origin$  $items$.
\textbf{The next step.}  Here we sort the separated items by their popularity, that is, by
the number of users who rated it. In table  (\ref{table:rating:matrix}) we give an example of
a rating matrix, where the popularity of item 1\_2 (item 1 with rating 2) in this matrix is 3.
 \begin{table}[htdp]
\begin{center}
\begin{tabular}{|c|c|c|c|c|c|}
\hline
	 & \textbf{item1} & \textbf{item2} & \textbf{item3} & \textbf{item4} & \textbf{item5}\\  \hline
 	\textbf{user1} & 2 &  &  & 1 & 1 \\  \hline
 	\textbf{user2} & 1 &  & 1 & 1 & 3 \\  \hline
 	\textbf{user3} & 2 & 1 & 2 & 3 &  \\  \hline
 	\textbf{user4} & 3 &  & 2 &  & 1 \\  \hline
 	\textbf{user5} &  & 3 & 3 & 1 & 1 \\  \hline
 	\textbf{user6} & 1 & 3 &  & 1 &  \\  \hline
 	\textbf{user7} & 1 &  & 1 & 2 & 3  \\  \hline
 	\textbf{user8} & 2 & 3 & 3 & 2 & 1 \\  \hline
	\textbf{user9} & 1 & 3 &  & 1 &  \\  \hline
	\textbf{user10} & 1 &  & 1 &  & 3 \\  \hline									
\end{tabular}
\end{center}
\caption{ \textit{Rating matrix, possible ratings [1, 2, 3] (k=3)}.}
\label{table:rating:matrix}
\end{table}

\textit{\textbf{The third step:}}
\label{par:ptd:third:step}
We adopt a topological representation by taking the sorted items set of
each domain and representing them by a forest of weighted graphs.
These graphs are constructed as follows:
	\textbf{Nodes}: Each item on the sorted items list is represented by one node (Note: the node represents a pair of an item and one rating value).
	\textbf{Edges}: An edge is found between two nodes if there are
	common users who rated both items represented by both nodes (Note: Rating must be the same as in the node).
	\textbf{Weight}: The weight of the edge between two nodes is defined as the Jaccard coefficient
	between the the sets of users who rated the items represented by the nodes.
	 Jaccard coefficient measures similarity between sample sets, and defined as the size of
	the intersection divided by the size of the union of the sample sets (Wikipedia):
	
	\textit  {Similarity(A, B) = J(A, B)} $ = \frac{|A \bigcap B|}{|A \bigcup B|} $. If the two sets $A$ and $B$ are empty then we
	define $J(A, B) = 0$. The max value of this similarity
	when $A, B$ are finite is 1, and the minimal value is 0.

\begin{figure}[h!t]
\centering
\includegraphics[width=2.9in, height=1.3in]{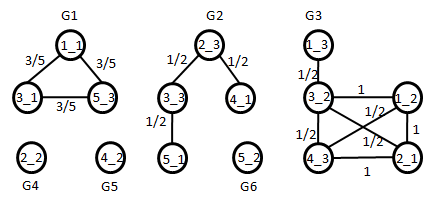}
\caption{\textit{Behavior graphs (forest) of table (\ref{table:rating:matrix}) when threshold = 0.5.}}
\label{fig:behavior:graphs}
\end{figure}

If the weight of edge is less than a given threshold \footnote{This threshold depends on different parameters
such as the number of users in the domain, the number of items, the number of ratings, etc. We set it as 0.5 in our experiments.}
we drop the edge. Note, a one-to-one mapping exists between the nodes and the items, thus we can use the term
node or item to refer the same element. Figure (\ref{fig:behavior:graphs}) shows the behavior
graphs (forest) that we get from table (\ref{table:rating:matrix}) when threshold = 0.5.

\textit{\textbf{The fourth step:}}
\label{par:ptd:fourth:step}
We convert each graph in the forest into a topological tree which we refer as \textit{a behavior tree}.
This is another representation of the users' behavior in the graph based on the graph's structure.
This process is done by adding an artificial node to be the root for each tree, and connecting all the popular
\footnote{The most popular items in each graph are those have greater popularity than the average popularities in the same graph.}
nodes in the graph to this node. Then we use a recursive greedy algorithm for completing the tree by adding all the other nodes on the graph to the tree.
This is done by connecting each child to the parent who is connected to the highest edge weight in the original graph.

\begin{figure}[h!t]
\centering
\includegraphics[width=3.4in, height=1.1in]{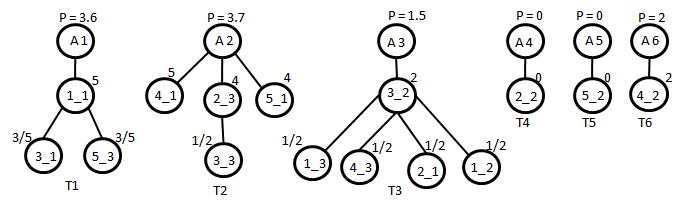}
\caption{\textit{Behavior trees constructed based on the behavior graphs (\ref{fig:behavior:graphs}). "P" is the popularity threshold.}}
\label{fig:behavior:tree}
\end{figure}

The motivation for moving from graphs to trees is the simpler hierarchical structure of the trees compared to graphs,
and the ability to represent sufficient data of the behavior graphs by behavior trees.
Figure (\ref{fig:behavior:graphs}) represents the graphs we get from the rating matrix in Table (\ref{table:rating:matrix}). and
Figure (\ref{fig:behavior:tree}) illustrates the behavior trees based on the behavior graphs
in figure (\ref{fig:behavior:graphs}), where $T_i$ is the behavior tree of graph $G_i$, $i: 1 \rightarrow 6$, and the nodes with labels A1-A6 are artificial nodes.

\subsubsection{Tree Matching}
\label{subsec:graph:matching}
The first task is to decide for each tree in the source domain which tree in the target domain is the most appropriate to be
matched with. Here we have the advantage of the common users, so we define similarity between
two trees in the same way that we defined similarity between two items in (\ref{par:ptd:third:step}), but here we refer to it on a set of users
instead of on a set of items:
\newline
	\textit  {Tree-Similarity(T1, T2) = J(U(T1), U(T2))}  $= \frac{|U(T1) \bigcap U(T2)|}{|U(T1) \bigcup U(T2)|}$
	%%, $J(A, B)$ is the same as in (\ref{par:ptd:third:step}), and\\
\newline
	\textit{U(T) = } $\bigcup _{i=1} ^{|T|}$ (\textit{u} $|$ user \textit{u} rated item \textit{i} in \textit{T}).
	
After we find for each tree in the source the tree in the target that it should be matched with, we run the tree matching process to match the
nodes in both trees. Eventually we arrive at a set of pairs of matched nodes. These pairs are the bridges between the items
in the source and the items in the target.
	
\subsubsection{Building the Training Samples}
\label{subsec:building:training:samples}
This is the final step in preparing training data. Here we prepare the training set for a supervised machine learning task. We find a special structure
for features vectors that combines the data of the users' behavior in the source and target domains, and the items content data.
This is done by taking all the matched nodes from the previous step. Then, for each two matched nodes $n_s$ and $n_{tr}$ that represent item $t_s$ in the source
with rate $r_s$ and item $t_{tr}$ in the target with rating $r_{tr}$ respectively, we define one features vector as following:\\
\textit{$[f_1(t_s),...,f_n(t_s), r_s ; f_1(t_{tr}), ...,f_m(t_{tr}), r_{tr} ]$}, where \textit{$f_i(t)$} is the value of feature $i$ for item $t$.
This structure provides a full image for the user's behavior from the previous step and the features of the items they rated.

\subsection{Model Training}
\label{sec:model:training}

Here we build a model of a multiclass classifier on the training samples which combine features of items in the source, ratings in the source, features of items in the target, and ratings in the target. Since our goal is to predict the user's rating of an item in the target domain, the class of the classifier is the last attribute in the features vector, where all the other attributes in the vector represent the features. We expect that this model to predict rating for an item \textit{t} in the target domain for a new user \textit{u}  that has rated only items in the source domain. The number of the classes equals the number of possible ratings values in the target domain.
\subsection{Recommendation}
\label{sec:recommendation}

The recommendation task is based on the model we described in the previous task (\ref{sec:model:training}).
When the system is asked to recommend the top $N$ items in the target domain for a new user \textit{u}, the system
ranks each item $tr$ in the target domain based on items that the user has rated in the source domain. The ranking is carried out
by building \textit{a features matrix} for each item $tr$ in the target as follows:
\begin{itemize}
	\item The number of the rows in the matrix equals the number of items the user rated in the source.
	\item The number of the column in the matrix equals  $n$ + $m$ + 2, where $n$ is the number of the item's features in the source, $m$ is the number of 	
	the item's features in the target domain and + 2 for the rating in the source and the class (rating in target).
	\item The value in row $i$ and column $j$ equals:
	\begin{description}	
		\item[\textbf{If }$j < n + 1$] Then: The value of feature $j$ in item $i$ (in the source)		
		\item[\textbf{If }$j = n + 1$] Then: The rating value that the user $u$ rated item $i$ (in the source)
		\item[\textbf{If }$j > n + 1$] Then: The value of feature $j-(n+1)$ of the item $tr$ (in the target)
		\item[\textbf{If }$j = n + 2$] Then: The value is "?" (missing)
	\end{description}
\end{itemize}

To find the rank of a features matrix $M$, we first run the classifier on each row in the matrix in order to return the vector of the predicted probability for each class.
This vector is called \textit{the distribution vector} and represents the probability for each class, so the value in entry $i$ equals to the probability that this samples is classified as $i$. Also, the vector's size equals the number of the classes (possible ratings). For each row we take the distribution vector $P$ returned by the classifier and compute the \textit{\textbf{expected rating}}\footnote{We assume that the ratings are natural continuos numbers that start from 1 as the minimal value.} as follows:
\textbf{\textit{expected rating}} = $\sum_{j=1}^k j \times P[j]$, when $k$ is the size of $V$.
The rank of the features matrix $M$ is defined as the average of the expected ratings of all the rows in $M$: $Rank(M) = \frac{\sum_{j=1}^z ER(j)}{z}$, where $ER(j)$ is the expected rating for row $j$ and $z$ is the number of the rows in $M$.
Then we sort these matrices by the rank value, and return the target items represented by the top $N$ matrices as the recommended items for the user $u$.

\section{Experiment}
\label{ch:experiment}
In this chapter we investigate whether the additional knowledge gained from the source domain and the content of the items can improve the
recommendation in the target domain. We compare our approach with the Popularity, which is a single domain approach
and usually used for recommendation to new users,
and with the KNN-cross-domain approach that uses both domains but dose not use
the content data.

\subsection {Dataset}
We used the Loads data-set in our experiments. Loads data is a real-world content dataset that includes different
domains such as videos, music, games (common users among the domains). We chose music-loads as the source domain and game-loads
as the target domain, and extracted 600 common users from both domains, 817 items from the music domain with 18,552 ratings, and 1264
items from the games domain with 17,640 ratings. This dataset is event-based, thus we manipulated it and converted the events to ratings by weighting
the events by their type. For example the event \textit{user buys an item} that was converted to the max rate value. For this experiment
we used a binary ratings.

\subsection{Evaluation and Metrics}
\label{sec:exp:eval:met}
Our goal is to recommend a set of $n$ items that may find the interest of the new user in the target domain. This kind of
recommendation refers to recommending some good items \cite{ecf04herkonterrie} or \textit{top-N} items.
The correct method of evaluating the recommender in this case is by measuring the
precision at $N$ (or Top-N precision) which is the number of interesting items from the recommended items \cite{ecf04herkonterrie}.
Since we have content data, we consider a recommended item as a true positive if it is similar to 80\% of the positively rated items
%that the user rated as positive.
\subsection{Baseline}
We compare our method, referred to as BGM (Behavior Graph Matching), with two base-line recommenders:

\textbf{KNN-cross-domain}: This recommender is a cross-domain recommender
based on collaborative filtering with Pearson's correlation coefficient method. The main idea behind the method
is to find the K-nearest-neighbors of the active user in the source domain who also rated items in the target domain, and to consider them as the
K-nearest-neighbors of the active user, who is also in the target domain, and then predict the user's ratings in the target domain based on the ratings
in the domain.

\textbf{Popularity}: This is a naive method that recommends the popular items to the new users.
This method is the simplest to implement and sometimes outperforms other algorithms \cite{cne10schpopung}
especially in a domain where most of the users have similar preferences.

\subsection{BGM versus Popularity}
We evaluate the two methods by 10-fold cross validation on the Loads dataset, when each fold includes 60 users for test and 540 users for train. The test set was considered as the new users' set in the target domain who have just ratings in the source domain. The behavior patterns were based on all the users in the source domain, and users belong to the training set in the target domain. For each user in the test set we asked the recommender to recommend the \textit{Top-N} items when \textit{N= 5, 10, 15, 20, 50, and 100}, and for each set of recommended items we measured the top-N precision per user. The popularity algorithm uses the training set in the target domain to find the popular item that was recommend for each of the test users, then we computed the top-N precision for each \textit{N}, and for each \textit{N} we found the average of the top-N precision for all of the users.

Figure (\ref{fig:results:charts}) shows the results, where we divided into 6 groups, each group representing a different \textit{N} value and containing
three columns, the first one representing the average \textit{Top-N} precision value for the BGM, and the second one representing the average of \textit{Top-N} precision value for the popularity algorithm. We note that BGM outperforms the popularity in most of the cases.
Thus, our conclusion, which is based on results, is that it is better to use the source domain to improve recommendation in the target domain.

\subsection{BGM Vs. KNN cross-domain}
Tis experiment was conducted in order to compare the performance of our approach with the collaborative-filtering cross-domain approach which also
uses the knowledge in the source domain to return recommendations to the target domain.
The main goal was to determine whether the BGM method makes maximum use of the content data of the items, and to check whether the proposed behavior patterns  work well for representing the users' behavior and transferring knowledge. As in the first experiment, this experiment, too, was conducted with a 10-fold cross validation on Loads dataset.

\begin{figure}[h!t]
\centering
\includegraphics[width=2.8in, height=1.30in]{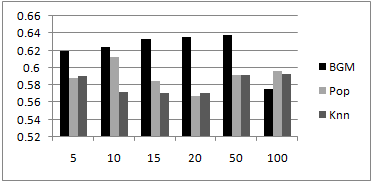}
\caption{\textit{Top-N precision values for BGM, Popularity, and Knn-cross-domain, with different N values}}
\label{fig:results:charts}
\end{figure}

We can compare the results in Figure (\ref{fig:results:charts}), where the third column in each group represents the value of top-N precision of the KNN-cross-domain.
Note: BGM outperforms the KNN-cross-domain in 83\% of the cases.

\subsection{Statistical Analysis}
We performed paired t-tests on the same results we get by running 10-fold cross-validation with BGM, Popularity and KNN-cross-domain on Loads dataset,
and we checked whether they were statistically significant. So each user was
considered a participant, and each of the recommenders was considered a different method of the test.

We received a table of 600 users (rows), and for each user we received the top-N precision value when \textit{TopN}= 5, 10, 15, and 20 with
each of the methods (12 columns per user). Then the t-test was performed for every two relevant columns (with the same \textit{N}) .
All the results are statistically significant, which means that our method performance is
significantly better than the two other methods when the number of recommended items is 5, 10, 15, and 20, and the size of the sample is 600 participants.

\section{Conclusions}
\label{ch:conclusion}
In this paper we presented and evaluated a novel method of transfer learning in content-based recommenders by using a new
topological structure that we called a behavior graph. The main idea of using such structure is its ability of representing rich data about
the users' behavior, and the relation between items in a structure. By employing tree matching methods we discovered a correlation between
items in the source and items in the target.

We compared our method with the popularity approach, which is generally used with recommendations to new users, and with
the KNN-cross-domain method. Our comparison was based on a real-world
dataset called Loads dataset. We evaluated the Top-N precision metric by 10-fold cross validation.

The results show that our method (referred to as BGM) outperforms the popularity and KNN-cross-domain methods in most cases.
Our conclusion: it is preferable to use the data in the source domain and the item's content data when dealing with this kind of
recommendation problem.

\bibliographystyle{abbrv}
\bibliography{thesisRef}  % sigproc.bib is the name of the Bibliography in this case

\end{document}